\documentclass[runningheads]{llncs}

\usepackage{eccv}

\usepackage{eccvabbrv}
\usepackage{multirow}
\usepackage{graphicx}
\usepackage{booktabs}
\usepackage{bm}
\usepackage[mathscr]{eucal}
\usepackage[accsupp]{axessibility}  
\usepackage[section]{placeins}
\usepackage[marginal]{footmisc}

\usepackage{hyperref}

\usepackage{orcidlink}
\makeatletter
\renewcommand*{\@fnsymbol}[1]{\ensuremath{\ifcase#1\or \dagger \or *\or \ddagger\or
   \mathsection\or \mathparagraph\or \|\or **\or \dagger\dagger
   \or \ddagger\ddagger \else\@ctrerr\fi}}
\makeatother

\begin{document}

\title{Thermal3D-GS: Physics-induced 3D Gaussians for Thermal Infrared Novel-view Synthesis} 

\titlerunning{Thermal3D-GS}

\author{Qian Chen\inst{1}\thanks{Co-first authors, contribute equally to this work} \and
Shihao Shu\inst{1,\dagger} \and
Xiangzhi Bai\inst{1,2,3,}\thanks{Corresponding author}
\orcidlink{0000-0002-6115-8237}}

\authorrunning{Qian Chen, Shihao Shu et al.}

\institute{Image Processing Center, Beihang University, Beijing, China \and
State Key Laboratory of Virtual Reality Technology and Systems,\\ Beihang University
\and
Advanced Innovation Center for Biomedical Engineering, Beihang University\\
\email{\{19376165, 19375326, jackybxz\}@buaa.edu.cn}}

\maketitle

\begin{abstract}
Novel-view synthesis based on visible light has been extensively studied. In comparison to visible light imaging, thermal infrared imaging offers the advantage of all-weather imaging and strong penetration, providing increased possibilities for reconstruction in nighttime and adverse weather scenarios. However, thermal infrared imaging is influenced by physical characteristics such as atmospheric transmission effects and thermal conduction, hindering the precise reconstruction of intricate details in thermal infrared scenes, manifesting as issues of floaters and indistinct edge features in synthesized images. To address these limitations, this paper introduces a physics-induced 3D Gaussian splatting method named Thermal3D-GS. Thermal3D-GS begins by modeling atmospheric transmission effects and thermal conduction in three-dimensional media using neural networks. Additionally, a temperature consistency constraint is incorporated into the optimization objective to enhance the reconstruction accuracy of thermal infrared images. Furthermore, to validate the effectiveness of our method, the first large-scale benchmark dataset for this field named Thermal Infrared Novel-view Synthesis Dataset (TI-NSD) is created. This dataset comprises 20 authentic thermal infrared video scenes, covering indoor, outdoor, and UAV(Unmanned Aerial Vehicle) scenarios, totaling 6,664 frames of thermal infrared image data. Based on this dataset, this paper experimentally verifies the effectiveness of Thermal3D-GS. The results indicate that our method outperforms the baseline method with a 3.03 dB improvement in PSNR and significantly addresses the issues of floaters and indistinct edge features present in the baseline method. Our dataset and codebase will be released in \href{https://github.com/mzzcdf/Thermal3DGS}{\textcolor{red}{Thermal3DGS}}.

\keywords{Thermal image \and Novel-view synthesis \and Physics-induced}
\end{abstract}

\section{Introduction}
\label{sec:intro}

Novel-view synthesis leverage image or sensor data and provide crucial support for applications such as virtual reality and autonomous driving. In recent years, novel-view synthesis in the visible light domain have been extensively studied. Implicit neural representation methods and 3D Gaussian splatting methods have both demonstrated excellent capabilities in synthesizing novel views. However, the image quality of visible light scenes is sensitive to weather conditions, with factors such as clouds and fog degrading image quality and leading to reconstruction failures. In contrast, thermal infrared images possess a unique all-weather imaging capability, unaffected by optical illumination and weather constraints. These characteristics enable the synthesis of novel views in scenes with strong environmental interference, such as mining tunnels, foggy environments, and nighttime scenes, highlighting the significant potential of thermal infrared novel-view synthesis tasks.

However, the physical characteristics of thermal infrared images also pose challenges for high-precision novel-view synthesis tasks. In thermal infrared imaging, atmospheric transmission effects exhibit variability with changes in viewpoint, giving rise to discrepancies in the characteristics of the same object across different perspectives. Consequently, synthesized novel viewpoint images may display color block disparities or floating artifacts. Additionally, thermal conduction between objects diminishes boundary information of different objects, causing issues such as blurred or disappearing edge features in novel viewpoint images. While these characteristics can be formalized and modeled using physical equations, the complexity of the imaging process often leads to the problem of multiple solutions in direct physical derivation.
 
To address the aforementioned challenges, this paper introduces a physics-induced 3D Gaussians splatting method, termed as Thermal3D-GS, for thermal infrared novel-view synthesis tasks. The proposed method leverages the unique physical characteristics of thermal infrared images and employs deep neural networks to simulate the corresponding parameters and physical processes of atmospheric transmission effects and thermal conduction. These parameters and processes are utilized to address the issues of floaters and indistinct edge features in synthesized images, respectively. Additionally, a temperature consistency constraint is introduced as a physical prior for objects in thermal infrared imaging, which is formalized and incorporated into the optimization objective as a loss function to enhance the network's performance.

To validate the effectiveness of the proposed approach, a large-scale Thermal Infrared Novel-view Synthesis Dataset (TI-NSD) is constructed. TI-NSD comprises 20 authentic thermal infrared scenes, including 7 indoor scenes, 7 ground-based outdoor scenes, and 6 UAV-based scenes, totaling 6,664 frames of thermal infrared images. To the best of our knowledge, TI-NSD stands as the world's first extensive dataset for thermal infrared novel-view synthesis task, providing a crucial benchmark for advancing research in the field.

Finally, extensive experiments are conducted on TI-NSD to evaluate the performance of novel-view synthesis methods, including Nerf-based methods and the 3D-GS method as baseline comparisons. The proposed Thermal3D-GS is validated to exhibit superior performance compared to the baseline algorithm 3D-GS. Across 20 scenes, Thermal3D-GS achieves an average improvement of 3.03 dB in PSNR over 3D-GS. Furthermore, Thermal3D-GS significantly outperforms comparative methods in terms of visual results. This underscores the substantial superiority of the proposed method in novel-view synthesis.

In summary, the contributions of this paper can be summarized as follows:

(1) A physics-induced 3D Gaussians splatting method, termed as Thermal3D-GS, is proposed specifically for thermal infrared novel-view synthesis tasks. Thermal3D-GS optimizes networks by modeling atmospheric transmission and thermal conduction physical processes through neural networks, and introduces a temperature consistency constraint as a loss function to optimize the network.

(2) The first large-scale dataset for thermal infrared novel-view synthesis named TI-NSD is established in this paper. TI-NSD comprises 20 authentic thermal infrared scenes, including 7 indoor scenes, 7 ground-based outdoor scenes, and 6 UAV-based scenes, totaling 6,664 frames of thermal infrared images.

(3) Extensive experiments are conducted on TI-NSD, evaluating the performance of mainstream methods on the proposed dataset. Additionally, through experiments, the proposed method demonstrated an average improvement of 3.03 dB in PSNR compared to baseline methods.

\section{Related Work}

\subsection{Novel-view Synthesis Methods}
Traditional approaches to scene reconstruction, exemplified by Structure-from-Motion (SFM)\cite{snavely2006photo} and Multi-View Stereo (MVS)\cite{hedman2018deep,eisemann2008floating,kopanas2021point,chaurasia2013depth}, involve re-projecting and blending original images to generate novel views, often yielding satisfactory results. Nevertheless, challenges arise when generating structures absent in the original images, leading to the production of artifacts. Additionally, storing all input images in the GPU incurs significant computational costs.

Neural Radiance Fields (NeRF)\cite{mildenhall2021nerf} have emerged as a pivotal method in the realm of novel-view synthesis, employing a neural network to model radiation at each 3D point in a scene, capturing light propagation and reflection. However, the use of large multi-layer perceptrons in NeRF significantly hampers rendering speed. Various strategies, such as grid-based\cite{wang2023f2,chen2022tensorf,sun2022direct,rho2023masked,fridovich2023k,cao2023hexplane} and hash-dependent\cite{muller2022instant,barron2023zip} approaches, adopt additional data structures to efficiently reduce MLP size and layers, resulting in improved inference speed. Noteworthy methods like InstantNGP\cite{muller2022instant} leverage hash grids for accelerated computations, while Plenoxels\cite{fridovich2022plenoxels} forsakes neural networks altogether, relying on sparse voxel grids for interpolating continuous density fields. Despite the improvement in rendering speed, these methods are still constrained by the need to query the light propagation and other steps, resulting in rendering speeds that remain unsatisfactory.

Recently, 3D-GS\cite{kerbl20233d} introduced an innovative technology centered on point cloud rendering. This technology attains superior rendering effects for novel-view synthesis, concurrently boasting a rendering speed of 100FPS. Its outstanding performance provides feasibility for new methods based on this technology, and the algorithmic improvements presented in this paper are also built upon 3D-GS.

\subsection{Tasks on Thermal Images}
Significant research is underway in the fields of detecting\cite{zhou2023lsnet,chatterjee2022breast,dai2021attentional}, tracking\cite{liu2020learning,lan2020modality}, and segmenting\cite{xiong2021mcnet,shivakumar2020pst900} thermal images. The Weighted Enhanced Local Contrast (WSLCM) \cite{han2020infrared} is introduced as a novel approach for thermal small target detection, utilizing local contrast calculation through differential weighting for the detection of weak and small targets. Additionally, \cite{li2020segmenting} leverages thermal image edge information to address blurred target boundaries and imaging noise, thereby enhancing the quality of thermal segmentation.

However, to the best of our knowledge, there is currently limited research on utilizing thermal images for  novel-view
synthesis. To address the additional geometric ambiguity arising from transparent objects or complex scenes, \cite{zhu2023multimodal} incorporates thermal images as supervision information to construct a multi-modal NeRF model. Nevertheless, this model solely employs thermal images as supervision and cannot exclusively output a novel-view image composed of thermal images.

\subsection{Novel-view Synthesis Datasets}
Mip-NeRF360 dataset\cite{barron2022mip}: The dataset comprises 9 scenes, with 5 outdoors and 4 indoors. Each scene features a complex central object or area alongside a detailed background, encompassing between 100 and 330 images.

Deep Blending dataset: \cite{hedman2018deep} collect 19 scenes. Each scene encompasses between 12 to 418 input images, featuring 5 indoor scenes and 5 scenes with a substantial amount of vegetation.

NeRF\_LLFF\cite{mildenhall2019local}: This dataset contains 8 real scenes produced by mobile phones, each scene has 20 to 62 frames.

A detailed comparison of the datasets is available in Table \ref{tb0}.
\begin{table}[]\centering
\caption{A detailed comparison of the datasets}
\label{tb0}
{\begin{tabular}{l|cccccc}
\hline
\multirow{2}{*}{Dataset}                                & \multirow{2}{*}{Indoor} & \multirow{2}{*}{Outdoor} & \multirow{2}{*}{UAV} & \multirow{2}{*}{Total} & \multicolumn{2}{c}{Frames} \\ \cline{6-7} 
                                                        &                         &                          &                      &                        & Min          & Max         \\ \hline
\begin{tabular}[c]{@{}l@{}}Mip-NeRF360\end{tabular}   & 4                       & 5                        & 0                    & 9                      & 100          & 330         \\ 
\begin{tabular}[c]{@{}l@{}}Deep Blending\end{tabular} & 5                       & 14                       & 0                    & 19                     & 12           & 418         \\ 
\begin{tabular}[c]{@{}l@{}}NeRF\_LLFF\end{tabular}     & 4                       & 4                        & 0                    & 8                      & 20           & 62          \\ 
\begin{tabular}[c]{@{}l@{}}TI-NSD (Ours)\end{tabular} & 7                       & 7                        & 6                    & 20                     & 132          & 492         \\ \hline
\end{tabular}}
\end{table}

\section{Thermal Infrared Novel-view Synthesis Dataset}
We collected a large-scale dataset named TI-NSD for domain of thermal infrared novel-view synthesis, comprising 20 distinct scenarios. Each scenario was captured in video format for approximately 90 seconds, yielding an excess of 50,000 frames of raw data. Novel-view synthesis data is extracted consistently at a fixed rate of 3 to 4 frames per second, with initial point cloud and camera pose acquisition estimated by colmap. Ultimately, we acquired a dataset comprising 6,664 images spanning 20 scenes, designated for both training and testing purposes.

Adhering to prevalent standards for novel-view synthesis datasets, we categorize these scenes into three types: indoor, outdoor, and UAV scenarios, including 7 indoor scenes, 7 ground-based outdoor scenes, and 6 UAV-based scenes. Indoor scenes encompass objects with varying temperature ranges, such as ice cola, hot water bottles, and stationary human models. Outdoor scenes depict objects under diverse lighting and weather conditions, including rain, snow, and sunny weather. The UAV scenes showcase different shooting angles, such as flat and overhead perspectives. 

For data collection, we utilized a Zenmuse XT thermal camera equipped with an uncooled vanadium oxide microbolometer. With a spectral range of 7.5-13.5$\mu m$, a pixel pitch of 17$\mu m$, and a focal length of 13$mm$, all images maintain a resolution of $480 \times 720$ pixels. 

To our knowledge, TI-NSD stands as the inaugural dataset applied to tasks related to thermal infrared novel-view synthesis. Refer to the supplemental material for detailed information regarding the dataset.

\section{Thermal3D-GS}
\label{sec:blind}
\renewcommand{\dblfloatpagefraction}{.9}

Leveraging 3D-GS, we generate 3D Gaussians, characterized by parameters such as position ($x$) and opacity ($\sigma$). Each 3D Gaussian incorporates spherical harmonics ($SH$)  to capture the view-dependent appearance. The ultimate 2D rendering is achieved employing a fast differentiable renderer, following iterative optimization of the 3D-GS parameters (such as position, covariance, and $SH$ coefficients) through adaptive control operations on the Gaussian density. However, this method does not consider the distinctive physical properties of thermal infrared radiation, resulting in floaters and blurry edges.

In this section, we commence with an analysis of the impact of imaging effects on thermal infrared novel-view synthesis, specifically addressing the manifestation of floaters and blurry edges in synthesized outputs. Subsequently, we present the framework of our proposed method, elucidating the strategies employed in leveraging imaging effects throughout this paper. Finally, a systematic discussion is provided on three thermal radiation effects: atmospheric transmission effects, thermal conduction, and temperature consistency constraints, detailing their modeling approaches and integration within the overarching framework.
\subsection{Motivation}

We assume a normalized thermal infrared 3D space, denoted as $X$, where 3D information of objects is solely determined by its actual thermal infrared radiation, unaffected by various possible influencing factors. The process of obtaining 2D images can thus be perceived as
\begin{equation}
\label{eq0}
 x_{t, \theta} = IM_{t, \theta}(f(X))
\end{equation}
where, $f(\cdot)$ represents the ideal process of projecting a 3D object into a 2D image without imaging process effects. $IM_{t, \theta}$ represents the influence of physical effects during the imaging process. As the physical effects vary with different angles $\theta$ and times $t$, $IM_{t, \theta}$ is a function of time $t$ and angle $\theta$.

The process of novel-view synthesis can be described as:
\begin{equation}
\label{eq0}
 \widetilde{X} = \mathop{\arg\min}\limits_{X}(\sum (f(X)-x_{t, \theta}))
\end{equation}

The imaging effect varies with different angles and times, resulting in each 
$x_{t, \theta}$ corresponding to a distinct 
$X_{t, \theta}$. Therefore, in the reconstruction process without considering the influence of imaging, 
$\widetilde{X}$ is essentially the maximum intersection of 
$X_{t, \theta}$, deviating significantly from the true 
$X$.

Once the imaging impact is taken into account, the novel-view synthesis is reformulated as:
\begin{equation}
\label{eq0}
 \widetilde{X}^{\prime} = \mathop{\arg\min}\limits_{X}(\sum f(X)-IM^{-1}_{t, \theta}(x_{t, \theta}))
\end{equation}
where, $\widetilde{X}^{\prime}$ can be considered the most accurate estimate of $X$.

\begin{figure}
    \centering
    \includegraphics[width=1\linewidth]{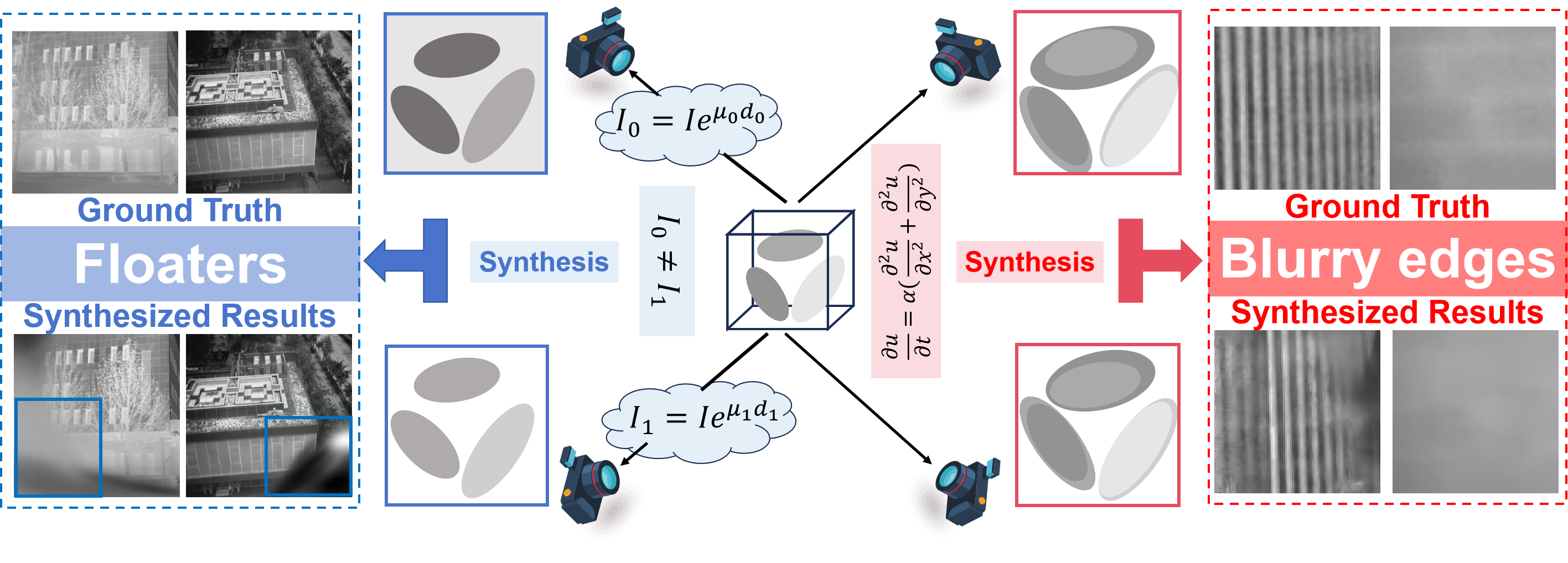}
    \caption{The imaging process is influenced by atmospheric transmission and thermal conduction. In the image, high-temperature objects exhibit brightness, while low-temperature objects appear dark. The left side primarily depicts the atmospheric transmission effect, resulting in directly synthesizing the novel-view image will result in floaters. The right side predominantly depicts the thermal conduction effect, leading to  directly synthesizing new perspective images will result in blurred edges.}
    \label{fig:4}
\end{figure}

This paper primarily examines the physical effects of atmospheric transmission and thermal conduction on thermal imaging, as depicted in Fig. \ref{fig:4}. 

\renewcommand\floatpagefraction{.9}

Atmospheric transmission effect: Atmospheric transmission effect leads to the attenuation of thermal radiation due to absorption and scattering, resulting in a notable decrease in the radiation intensity of thermal infrared images compared to the radiation intensity of actual 3D objects. The decline varies with the imaging angle, indicating that the imaging results of the same object from multiple viewing angles differ. During the optimization of 3D-GS, a relatively balanced $\widetilde{X}$ is learned, and the $\widetilde{x}$ projected by this $\widetilde{X}$ differs to some extent from the real $x$. To compensate for this difference, 3D-GS tends to learn $floaters_{t, \theta}$, ensuring:
\begin{equation}
IM_{t, \theta}(f(\widetilde{X}+floaters_{t, \theta}))=x
\end{equation}
where, $floaters_{t, \theta}$ represents the incorrectly learned floaters.

Thermal conduction: High-temperature objects exhibit pronounced thermal conduction, heating the surrounding medium and leading to significant artifacts, while low-temperature objects show less noticeable thermal conduction less severe artifacts. The radiation intensity of the artifact diminishes gradually with distance from the object, frequently manifesting as the object's edge in the image. Artifact edges resulting from the thermal conduction exhibit variations across multi-view thermal infrared images. Throughout the optimization process, akin to the concept of multi-frame averaging, an averaged artifact edge $\widetilde{X}$ is acquired, often leading to the manifestation of edge blurring $\widetilde{x}$.

\subsection{Overall Framework}
\begin{figure*}
    \centering
    \includegraphics[width=1\linewidth]{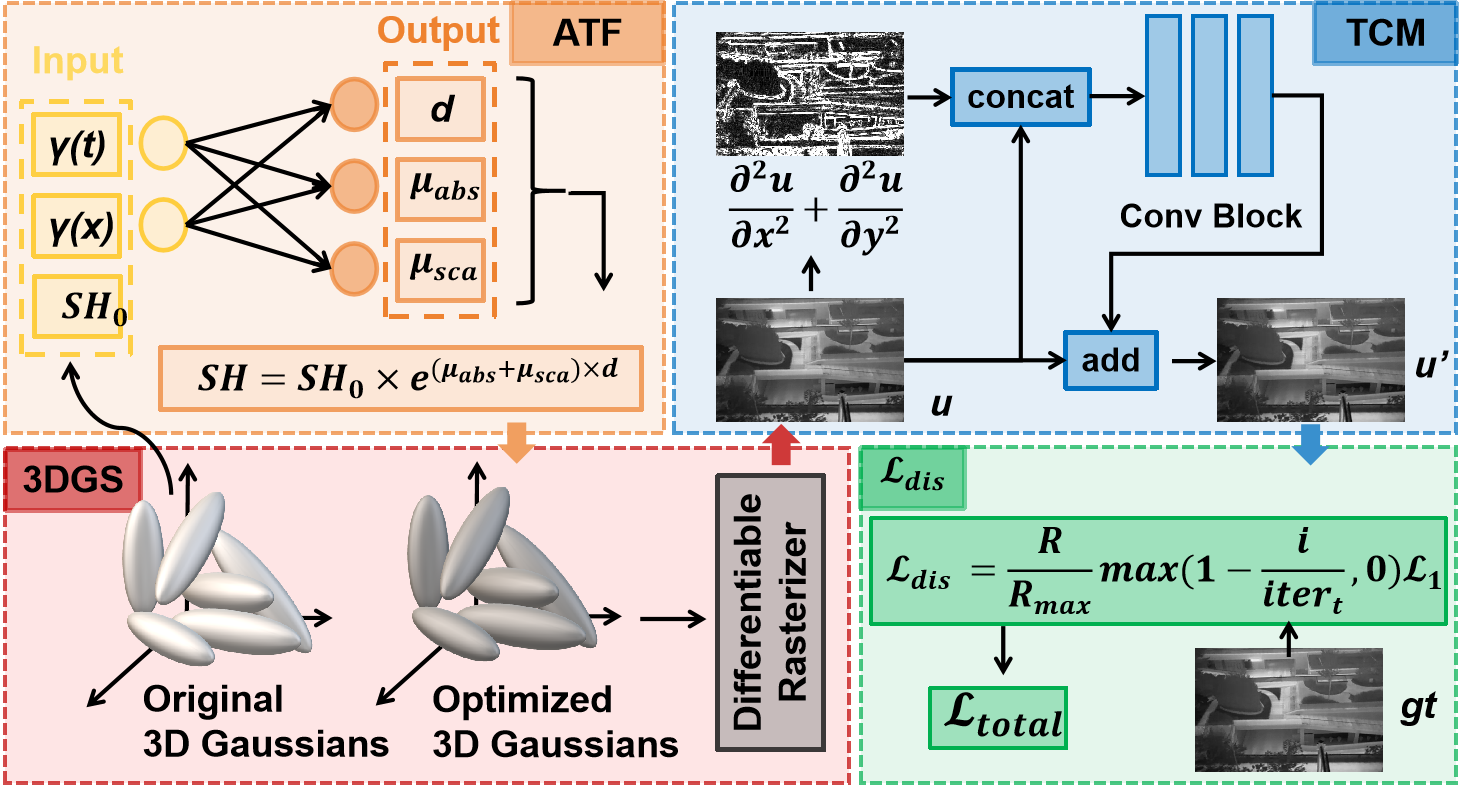}
    \caption{Overview of the proposed framework.}
    \label{fig:1}
\end{figure*}

Drawing upon the aforementioned description of the imaging process, this paper presents the overarching framework of Thermal3D-GS. As illustrated in Fig. \ref{fig:1}, this study extracts the original $t$, $x$, and $SH $ synthesized by 3D-GS. Atmospheric Transmission Field (ATF) is developed to optimize the $SH$ for the fibrillar structures present in the synthesized images, yielding the optimized Gaussians. Subsequently, Thermal Conduction Module (TCM) is devised to further refine the synthesized images, specifically targeting blurry edges. Furthermore, this paper introduces a temperature consistency loss to constrain the network, thereby enhancing its sensitivity and robustness to irregular regions. The physical underpinnings and modeling methodologies of each module will be elucidated in detail in the following three subsections.

\subsection{Atmospheric Transmission Field}

During atmospheric transmission, thermal radiation undergoes influences from various molecules and particles. Greenhouse gases, like water vapor and carbon dioxide, exhibit strong absorption capabilities for thermal infrared radiation. Additionally, nitrogen, oxygen molecules, and cloud particles contribute to the scattering of thermal radiation in the atmosphere. Consequently, thermal radiation experiences attenuation after traversing the atmosphere, a phenomenon described by the Bouguer-Lambert-Beer law\cite{vollmer2021infrared}:
\begin{equation}
\label{eq1}
 I = I_{0} e^{\mu(\lambda)d}
\end{equation}
In the formula, $I_{0}$ denotes the initial radiation intensity at a propagation distance of 0; $\mu = \mu_{abs} + \mu_{sca}$ represents the medium attenuation coefficient, consisting of absorption and scattering components respectively; and $d $ signifies the propagation distance.

The absorption and scattering of thermal radiation in the atmosphere are affected by temperature, humidity, etc. We assume that each 3D Gaussian represents a small continuous area in space, so they share the uniform attenuation coefficient($\mu_{abs}$, $\mu_{sca}$ and $d$). Simultaneously, we make the assumption that the attenuation of thermal radiation intensity corresponds solely to the reduction in grayscale within the thermal infrared image.

We decouple the effects of attenuation and geometry using an MLP network (ATF), enabling the independent learning of attenuation-free geometry and attenuation coefficients. This approach effectively mitigates the impact of atmospheric transmission on thermal infrared. More specifically, outlined by the yellow box in Fig. \ref{fig:1},  the ATF network takes the positional encoded 3D Gaussian position $\gamma(x)$ and the shooting time $\gamma(t)$ as input to determine the attenuation coefficient of the 3D Gaussian at that particular moment:
\begin{equation}
\label{eq2}
(\mu_{abs}, \mu_{sca}, d) = \mathscr{F}_{ATF}(\gamma(x),\gamma(t))
\end{equation}

In typical novel-view synthesis datasets, the time interval between each frame remains constant. Consequently, we employ a natural sequence of 1, 2, and so forth, to denote the normalized shooting time. $\gamma$ denotes the positional encoding\cite{mildenhall2021nerf}:
\begin{equation}
\label{eq3}
\gamma(p)=(sin(2^{k}\pi p), cos(2^{k}\pi p))_{k=0}^{L-1}
\end{equation}
where $L$ is the number of the frequencies, and $L = 10$ for both $x$ and $t$. 

This obtained coefficient is then applied to the normalized radiation intensity of each 3D Gaussian, resulting in the authentic radiation intensity after attenuation:
\begin{equation}
\label{eq4}
SH = SH_{0} e^{(\mu_{abs} + \mu_{sca})d}
\end{equation}
Here, SH denotes the coefficient of the spherical harmonic function of this 3D Gaussian. In visible light imaging, its attenuation corresponds to the attenuation of light at that location; in thermal infrared imaging, its attenuation equates to the attenuation of thermal radiation at that location.

We set depth of ATF network $D = 8$ and the dimmension of hidden layer $W = 256$. At the same time, we set the initial parameter $\mu_{abs}=\mu_{sca}=0$ and $d=1$.
\subsection{Thermal Conduction Module}

Thermal conduction is a ubiquitous phenomenon facilitating the transfer of thermal energy through molecular vibrations and collisions within a substance. Whether in a solid, liquid, or gas, thermal conduction manifests in the presence of a temperature difference. In this process, energy moves from high-energy molecules in the high-temperature region to low-energy molecules in the low-temperature region, achieving thermal equilibrium. The rate of heat transfer hinges on the thermal conductivity and temperature gradient of the material. In thermal infrared imaging, thermal conduction among distinct objects can result in indistinct object boundaries. By incorporating thermal conduction modeling, the energy transfer between objects can be comprehensively grasped, enhancing sensitivity to temperature changes and improving novel-view synthesis accuracy.

Within the 2D temperature field, a specific rectangular coordinate system and microelement $dS=(dx, dy)$ are adopted. The temperature, denoted as $u=u(t,x,y)$, at each point and time $t$ serves as the representative quantity for thermal motion.

The heat flow adheres to Fourier's law of thermal conduction\cite{einstein1905molekularkinetischen}, where heat moves from high to low temperatures in a specific direction. The quantity of heat moving in a given direction is proportionate to the rate of temperature decrease in that direction. The mathematical expression for this phenomenon is:
\begin{equation}
\label{eq5}
\bm{Q_{n}}=-k(x, y; \bm{n})\frac{\partial{u}}{\partial{n}}\bm{n}
\end{equation}

where $\bm{Q_{n}}$ represents the heat flow density vector in the $\bm{{n}}$ direction, signifying the heat passing through the unit area in the $\bm{{n}}$ direction per unit time. The term $k (x, y; \bm{n}) $ denotes the thermal conduction coefficient of the medium, maintained as a constant under the assumption of a uniform and isotropic medium, and is denoted as $k$.

The heat introduced through x direction of the microelement during the time interval $[t, t + dt]$ is:
\begin{equation}
\begin{split}
\begin{aligned}
\label{eq6}
Q_{X}
&=-k\frac{\partial{u}}{\partial{x}}|_{(t,x,y)}dtdy+k\frac{\partial{u}}{\partial{x}}|_{(t,x+dx,y)}dtdy \\ 
&\approx k\frac{\partial^{2}{u}}{\partial{x^{2}}}dtdxdy
\end{aligned}
\end{split}
\end{equation}

In the same way, the heat in y direction can be obtained as:
\begin{equation}
\label{eq7}
Q_{y}=k\frac{\partial^{2}{u}}{\partial{y^{2}}}dtdxdy
\end{equation}

The heat required to increase the temperature of the microelement is:
\begin{equation}
\label{eq8}
c \rho [u(t+dt, x, y)-u(t, x, y)]dxdy \approx c \rho \frac{\partial{u}}{\partial{t}}dtdxdy
\end{equation}
where $c$ and $\rho$ are specific heat capacity at constant pressure and density respectively.

In accordance with the law of conservation of energy, the heat needed to elevate an object's temperature equals the sum of externally incoming heat and internally generated heat: 
\begin{equation}
\label{eq9}
k\frac{\partial^{2}{u}}{\partial{x^{2}}}dtdxdy+k\frac{\partial^{2}{u}}{\partial{y^{2}}}dtdxdy=c \rho \frac{\partial{u}}{\partial{t}}dtdxdy
\end{equation}

Thus, the 2D temperature field thermal conduction formula is obtained:
\begin{equation}
\label{eq10}
\frac{\partial{u}}{\partial{t}}=\alpha\mit\Delta u
\end{equation}
where $\mit\Delta=\frac{\partial^{2}}{\partial{x^{2}}} + \frac{\partial^{2}}{\partial{y^{2}}}$ is 2D Laplacian
operator, $\alpha = \frac{k}{c \rho}$ is a constant that reflects the response speed of a substance to thermal conduction.

As the thermal infrared image represents a 2D temperature field, Eq. \ref{eq10} is applicable to it as well. This equation indicates that the impact of thermal conduction on thermal imaging is governed by the constant $\alpha$ and the second differential of the thermal image.

In consideration of the pixel-wise heterogeneity of $\alpha$, traditional methods rooted in physics encounter challenges in accurately modeling the impact of various factors. Hence, this paper introduces a TCM based on deep learning to effectively capture and model the intricate relationship between the original infrared image and the second-order gradient image, mimicking the intricate thermal characteristics. The TCM, outlined by the blue box in Fig. \ref{fig:1}, first extracts the second-order gradient features of the input image and then utilizes convolutional blocks to fuse the input image with gradient information. This is employed to simulate $\alpha$ at different pixel positions and incorporates them into the reconstruction process through a residual addition mechanism to address thermal losses induced by thermal conduction. We set the depth of convolutional layers in TCM as $D=3$, input feature dimensions as $W_{i}=[2n,n,n]$, and output feature dimensions as $W_{o}=[n,n,n]$, where $n$ represents the input image feature dimensions.

\subsection{Discontinuous Loss}
In real-world scenarios, the temperature of object surface typically exhibits smooth and continuous changes, with few abrupt variations. This smooth transition signifies a relatively uniform heat distribution inside and on the object's surface. In thermal images, this smoothness is often reflected in a gradual grayscale distribution, presenting few corners. Consequently, the appearance of corners in the image is more likely attributed to errors in model learning. To address this, this paper introduced the discontinuous loss, aiming to prompt the model to allocate more attention to regions with corner points. These areas indicate potential issues or abnormalities in the image, thereby enhancing the model's sensitivity to irregular situations and fortifying its robustness.

In the context of the Harris corner detection algorithm\cite{harris1988combined}, the formulation for constructing the corner response function R is as follows: 
\begin{equation}
\label{eq11}
 R=det(M) - k(trace(M))^{2} 
 \end{equation}
Here, $M$ symbolizes the covariance matrix of the image, $det(\cdot)$ denotes the determinant of the matrix, and $trace(\cdot)$ signifies the trace of the matrix.

Hence, utilizing the corresponding function of the corner point, the discontinuous loss is formulated as follows:
\begin{equation}
\label{eq12}
\mathcal{L}_{dis}=\frac{R}{R_{max}} max(1-\frac{i}{iter_{t}}, 0) \mathcal{L}_{1}
 \end{equation}

 In the Formula, the first term represents the normalized corner point response, signifying the likelihood of the pixel being a corner point. The second term serves as an attenuation factor associated with the number of training iteration, here, the $iter_{t}$ is set to 5,000. The third term is the absolute loss, expressing the absolute distance between the generated image and the ground truth.

Therefore, the ultimate loss is formulated as follows, and we use $\lambda_{dis}=\lambda=0.2$ in all of our tests.
\begin{equation}
\label{eq13}
\mathcal{L}_{total}=\lambda_{dis}\mathcal{L}_{dis}+\lambda\mathcal{L}_{D-SSIM}+(1-\lambda_{dis}-\lambda)\mathcal{L}_{1}
 \end{equation}

\section{Experiments}
\renewcommand{\dblfloatpagefraction}{.9}
\begin{figure*}
    \centering
    \includegraphics[width=1\linewidth]{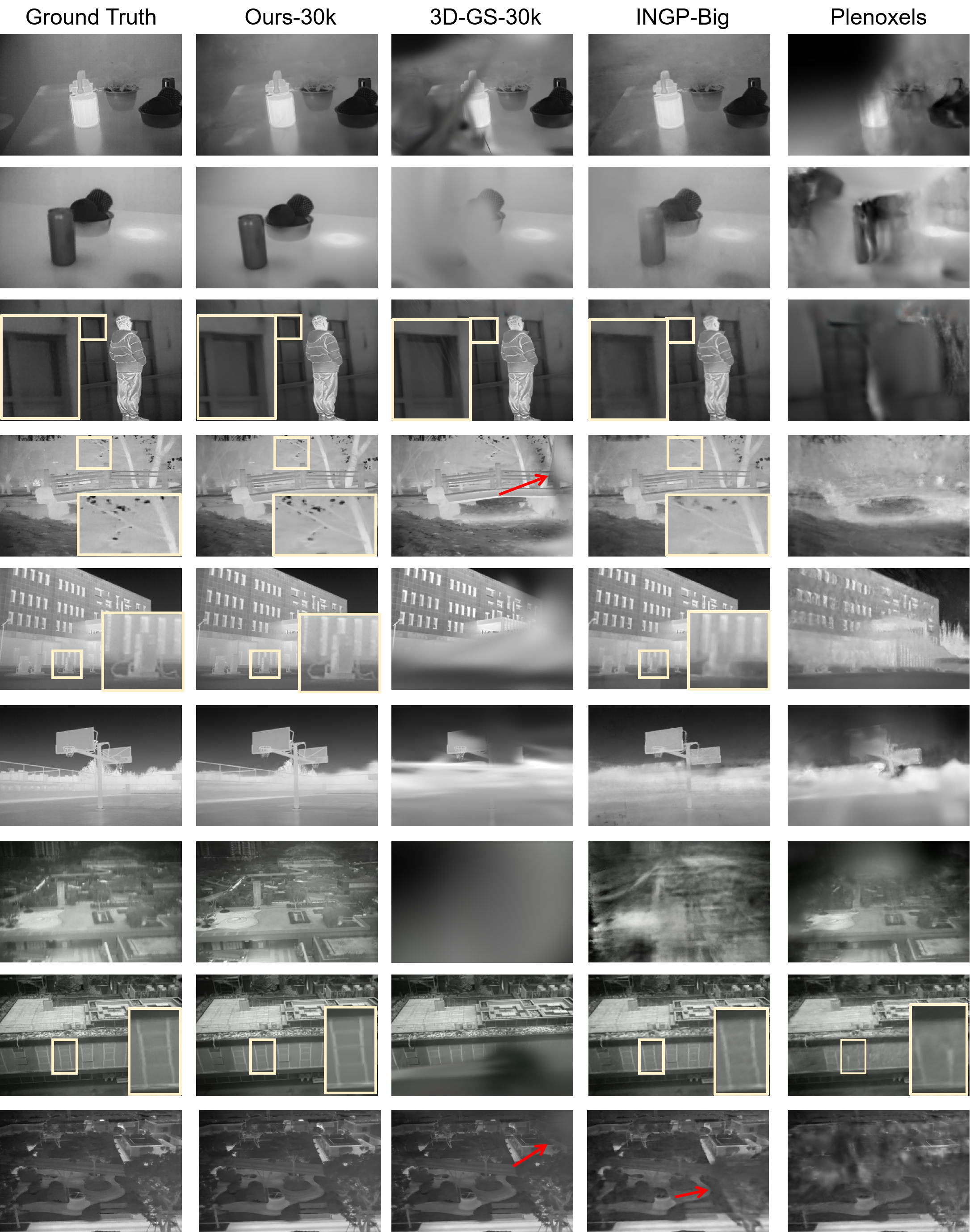}
    \caption{We show comparisons of ours to previous methods and the corresponding ground truth images from held-out test views. The scenes are, from the top down: Merge, Apples, Standing, Bridge, Tall Building, Basketball Court, UAV1, UAV2, UAV5. Non-obvious differences in quality highlighted by arrows/insets.}
    \label{fig:3}
\end{figure*}
Our ATF and TCM were implemented using PyTorch \cite{paszke2019pytorch}, and all experiments were conducted on a 3090 with 24GB of memory.
\subsection{Implementation Details}
We retained the differential Gaussian rasterization proposed by 3D-GS \cite{kerbl20233d}. The training involved a total of 30,000 iterations. Optimization utilized a single Adam optimizer \cite{kingma2014adam}, but each component had a different learning rate: the 3D Gaussians maintained the same learning rate as the official implementation, the TCM module shared the learning rate with the 3D Gaussians, and the ATF network experienced exponential decay in learning rate, ranging from $8\times10^{-4}$ to $1.6\times10^{-6}$. The beta values for Adam were set to (0.9, 0.999).

\begin{table*}[]
\centering
\scriptsize
\caption{Quantitative evaluation of our method compared to previous work.}
\label{tb2}
\begin{tabular}{l|lll|lll|lll|lll}
Scene     & \multicolumn{3}{c|}{Indoor}                                                                   & \multicolumn{3}{c|}{Outdoor}                                                                  & \multicolumn{3}{c|}{UAV}                                                                      & \multicolumn{3}{c}{Average}                                                                   \\
Method    & SSIM                          & PSNR                          & LPIPS                       & SSIM                          & PSNR                          & LPIPS                         & SSIM                          & PSNR                          & LPIPS                         & SSIM                          & PSNR                          & LPIPS                         \\ \hline
Plenoxels & 0.867                         & 22.13                         & 0.385                         & 0.768                         & 22.15                         & 0.433                         & 0.780                         & 25.56                         & 0.351                         & 0.805                         & 23.28                         & 0.390                         \\
INGP-Base & 0.916                         & 26.99                         & 0.291                         & 0.837                         & 26.00                         & 0.302                         & 0.681                         & 20.86                         & 0.404                         & 0.811                         & 24.62                         & 0.332                         \\
INGP-Big  & 0.918                         & 27.46                         & 0.289                         & 0.839                         & 26.45                         & 0.292                         & 0.680                         & 20.82                         & 0.387                         & 0.812                         & 24.91                         & 0.323                         \\
3D-GS-7k   & 0.951                         & 31.28                         & 0.270                         & 0.892                         & 26.38                         & 0.250                         & 0.940                         & 31.52                         & 0.146                         & 0.927                         & 29.64                         & 0.226                         \\
3D-GS-30k  & 0.953                         & 32.98 & 0.259 & 0.904                         & 28.89 & 0.227                         & 0.953 & 34.51 & 0.119 & 0.936                         & 32.01 & 0.206 \\
Ours-7k   & 0.957 & 32.69                         & 0.265                         & 0.922 & 28.33                         & 0.221 & 0.952                         & 33.60                         & 0.135                         & 0.943 & 31.44                         & 0.211                         \\
Ours-30k  & \bf{0.962} & \bf{36.01} & \bf{0.252} & \bf{0.942} & \bf{32.60} & \bf{0.187} & \bf{0.962} & \bf{36.74} & \bf{0.112} & \bf{0.955} & \bf{35.04} & \bf{0.187}
\end{tabular}
\end{table*}
\subsection{Results and Comparisons}

We use a train/test split, every 8th photo is selected for the test set to ensure consistent and meaningful comparisons for generating error metrics. We utilize standard metrics such as PSNR, L-PIPS, and SSIM, employed in the literature. 

We present results obtained by running InstantNGP in the base configuration (Base) and a slightly larger network recommended by the authors (Big), and two configurations of 3D-GS at 7K and 30K iterations, refer to Table \ref{tb2} for details. Here, the average results are presented, with the detailed experimental results for all scenes available in the supplemental materials.

Fig. \ref{fig:3} illustrates the differences in visual quality. Visual analysis indicates that 3D-GS tends to generate erroneous 3D Gaussians, manifesting as small floats and blurred edges in the image(Scene Merge, Scene  Apples, Scene UAV1); the outcomes produced by InstantNGP exhibit relative blurriness and lack fine details(Scene Standing, Scene Bridge, Scene Basketball Court); Plenoxels exhibit severe floaters and are not suitable for thermal infrared images. It is noteworthy to highlight that concerning UAV1, owing to the higher velocity of UAV during flight resulting in motion blur, solely our proposed technique demonstrates superior scene reconstruction capabilities, further excelling in the mitigation of motion blur evident in the ground truth imagery.

Furthermore, we conducted comparative experiments with ThermoNeRF, a method for new perspective synthesis using dual modalities of infrared and visible light. Details are discussed in the supplementary materials.

 \renewcommand{\dblfloatpagefraction}{.9}
\begin{figure*}
    \centering
    \includegraphics[width=1\linewidth]{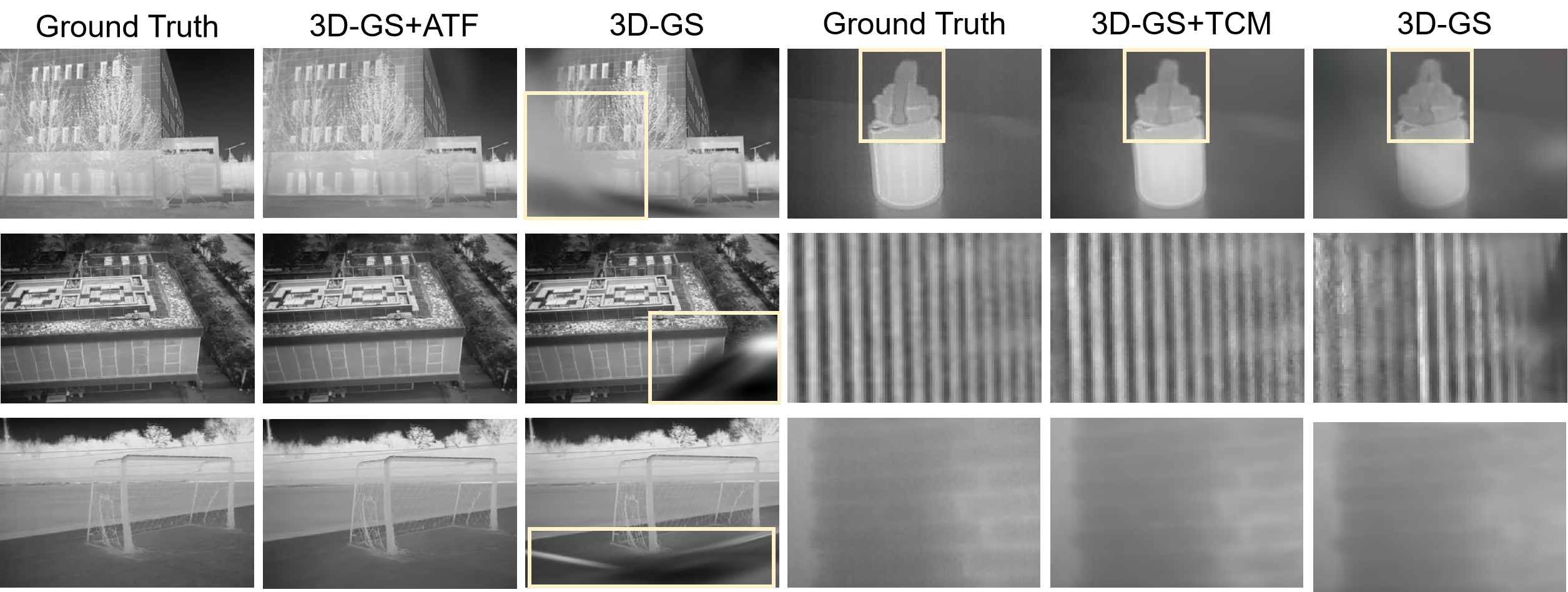}
    \caption{Visualization of ablation study. On the left, ATF eradicates floaters in the synthesized novel-view image. On the right, TCM effectively enhances blurred edges.}
    \label{fig:Ablation}
\end{figure*}

\subsection{Ablation study}
In Table \ref{tb3} we present an ablation study of our model on the TI-NSD. The average results were presented here and the detailed ablation study results for all 20 scenes are available in the supplementary materials.

To provide a clearer demonstration of each module's function, we display the visualization results depicted in the Fig. \ref{fig:Ablation}. Through the visualization, it becomes evident that the incorporation of ATF into 3D-GS notably diminishes floaters resulting from atmospheric transmission, while TCM adeptly sharpens blurred edges through the simulation of a 2D thermal conduction formula.

\begin{table*}[]
\centering
\scriptsize
\caption{Results for ablation study.}
\label{tb3}
\begin{tabular}{l|lll|lll|lll|lll}

Scene             & \multicolumn{3}{c|}{Indoor} & \multicolumn{3}{c|}{Outdoor} & \multicolumn{3}{c|}{UAV} & \multicolumn{3}{c}{Average} \\
Method            & SSIM   & PSNR   & LPIPS   & SSIM     & PSNR    & LPIPS   & SSIM   & PSNR   & LPIPS  & SSIM    & PSNR    & LPIPS   \\ \hline
3D-GS              & 0.953   & 32.98   & 0.259   & 0.904    & 28.89   & 0.227   & 0.953  & 34.51  & 0.119  & 0.936   & 32.01   & 0.206   \\
3D-GS+ATF          & 0.958   & 35.12   & 0.254   & 0.934    & 31.53   & 0.197   & 0.962  & 36.65  &\bf{0.110}  & 0.951   & 34.33   & 0.191   \\
3D-GS+TCM          & 0.955   & 33.72   & 0.257   & 0.916    & 29.68   & 0.197   & 0.955  & 35.04  & 0.118  & 0.941   & 32.70   & 0.194   \\
3D-GS+$\mathcal{L}_{dis}$        & 0.858   & 33.68   & 0.353   & 0.918    & 29.91   & 0.211   & 0.957  & 35.01  & 0.116  & 0.909   & 32.76   & 0.232   \\
Ours & \bf{0.962}   & \bf{36.01}   & \bf{0.252}   & \bf{0.942}    & \bf{32.60}   & \bf{0.187}   & \bf{0.962}  & \bf{36.74}  & 0.112  & \bf{0.955}   & \bf{35.04}   & \bf{0.187}  
\end{tabular}
\end{table*}

\section{Conclusion}
This paper elucidates the influence of thermal infrared physical characteristics on thermal infrared novel-view synthesis and introduces Thermal3D-GS from a physical perspective. Thermal3D-GS addresses the issues of floaters and indistinct edge features caused by atmospheric transmission and thermal conduction effects by separately modeling these two physical processes and fitting them using neural networks to refine the results of novel-view synthesis. Additionally, a loss function based on object surface radiation consistency characteristics is constructed for iterative optimization of Thermal3D-GS.

To validate the effectiveness of Thermal3D-GS, this paper proposes the first large-scale thermal infrared novel-view synthesis dataset, TI-NSD, comprising 20 scenes and 6,664 frames of authentic thermal infrared images. The dataset includes 7 indoor, 7 outdoor, and 6 UAV scenes, providing a crucial data foundation for research in this domain.

Extensive validation of Thermal3D-GS is conducted based on TI-NSD, demonstrating its efficacy across all 20 scenes. Compared to the baseline method 3D-GS, Thermal3D-GS achieves a notable 3.03 dB PSNR improvement and exhibits significant advantages in visual comparisons, substantially alleviating issues of floaters and indistinct edge features. The experimental results underscore the importance of integrating infrared physical characteristics in thermal infrared novel-view synthesis, highlighting the pioneering insights offered by the proposed method for advancing research in this field.


\section*{Acknowledgements}
This work was supported by the National Natural Science Foundation of China under Grant 62271016, the Beijing Natural Science Foundation under Grant 4222007, and the Fundamental Research Funds for the Central Universities.

%
%
\bibliographystyle{splncs04}
\bibliography{main}
\end{document}